\documentclass{article}
\usepackage{spconf,amsmath,graphicx}
\usepackage{listings}
\usepackage{ragged2e}
\usepackage{setspace}

\graphicspath{
{./figures/}
}

\title{BATCH-NORMALIZED RECURRENT HIGHWAY NETWORKS}
\name{Chi Zhang, 
      Thang Nguyen,
      Shagan Sah,
      Raymond Ptucha,
      Alexander Loui$^{\dagger}$,
      Carl Salvaggio
}
\address{ Rochester Institute of Technology, Rochester, NY 14623, USA \\
    $^{\dagger}$ Kodak Alaris Imaging Science R\&D, Rochester, NY 14615, USA} 
\begin{document}
%
\maketitle
\begin{abstract}
Gradient control plays an important role in feed-forward networks applied to various computer vision tasks. Previous work has shown that Recurrent Highway Networks minimize the problem of vanishing or exploding gradients. They achieve this by setting the eigenvalues of the temporal Jacobian to $1$ across the time steps. In this work, batch normalized recurrent highway networks are proposed to control the gradient flow in an improved way for network convergence. Specifically, the introduced model can be formed by batch normalizing the inputs at each recurrence loop. The proposed model is tested on an image captioning task using MSCOCO dataset. Experimental results indicate that the batch normalized recurrent highway networks converge faster and performs better compared with the traditional LSTM and RHN based models. 
\end{abstract}
\begin{keywords}
Gradient control, recurrent highway network, batch normalization, vanishing gradient, exploding gradient
\end{keywords}
\section{Introduction}
\label{sec:intro}

Recent progress in deep learning using Convolutional Neural Networks (CNNs) has achieved remarkable performance on various pattern recognition tasks. Increasing the depth of the networks significantly reduces the error on competitive benchmarks \cite{He2016}. However, training very deep networks is challenging due to the fact that the distribution of each layer`s inputs changes during training. Moreover, when training Recurrent Neural Networks (RNNs), gradients are unstable, and can vanish or explode over time. 

Several techniques \cite{Hochreiter1997,Srivastava2015,Zilly2016} have been proposed to circumvent the vanishing and exploding gradient problem. Batch normalization \cite{Ioffe2015} addresses the internal covariate shift problem by normalizing the layer inputs per mini-batch.  This speeds up training by allowing the usage of more aggressive learning rates, creates more stable models which are not as susceptible to parameter initialization, and has been shown to minimize vanishing and exploding gradients. While batch normalization has been found to be very effective for feedforward CNNs, the technique has not been as prevalent on RNNs. Laurent \textit{et al.} \cite{Laurent2016} reported that applying batch normalization to the input-to-hidden transitions of RNNs leads faster convergence but does not seem to improve the generalization performance on sequence modeling tasks. Cooijmans \textit{et al.} \cite{Cooijmans2016} found that it is both possible and beneficial to batch normalize both the input-to-hidden and hidden-to-hidden transition, thereby reducing internal covariate shift between time steps.

\begin{figure}
  \centering
  \centerline{\includegraphics[width=.45\textwidth,trim=3 2 0 0,clip]{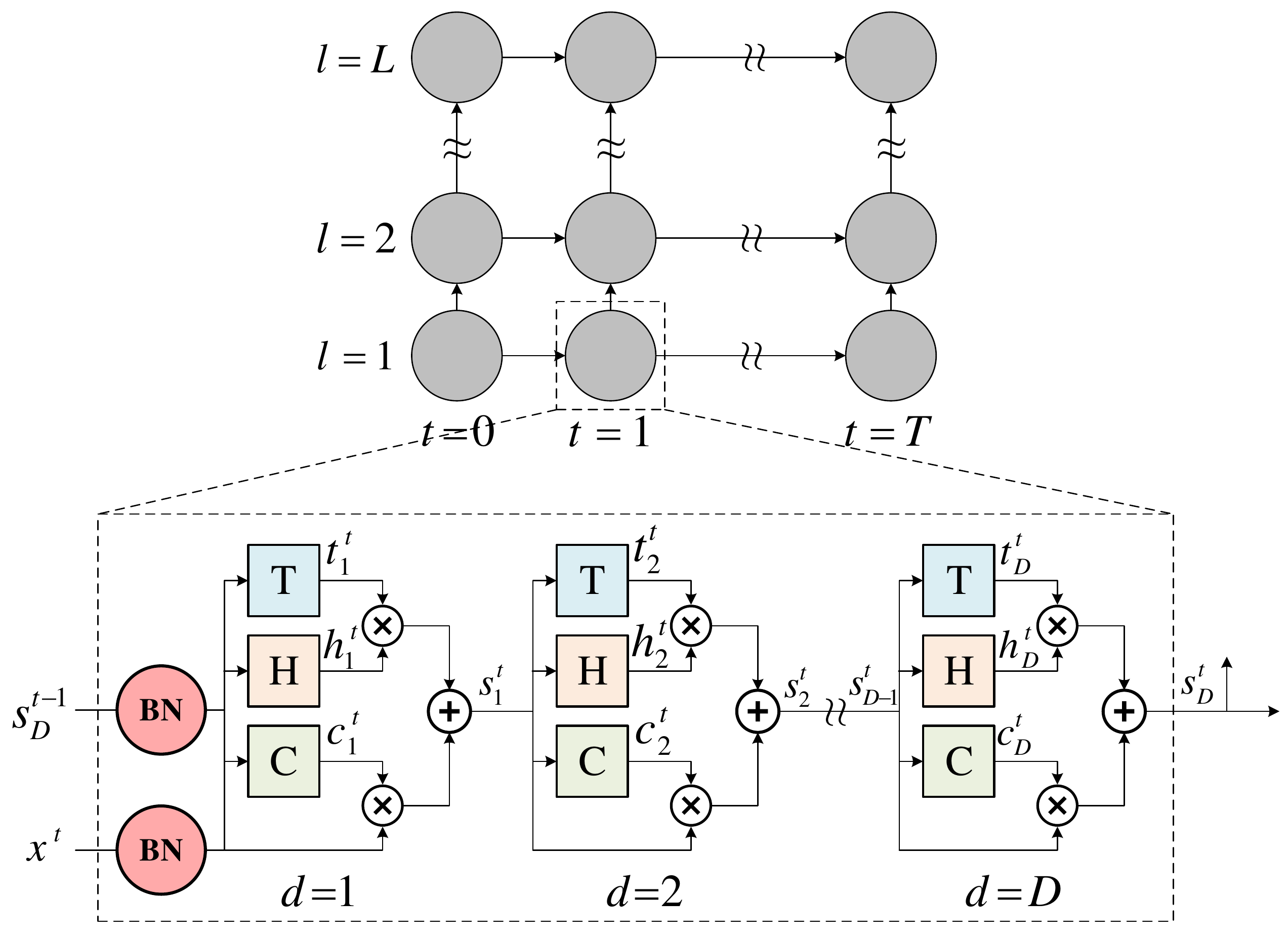}}
\caption{The architecture of batch normalized recurrent neural networks. T and C denote transform and carry gates specified in Equation \ref{eq:2} and \ref{eq:3} respectively, H is a nonlinear transform specified by Equation \ref{eq:1}, and BN represents batch normalization operation.}
\label{fig:archi}
\end{figure}

In addition to batch normalization, much attention has been paid to controlling gradient behavior by changing the network structure. For instance, networks with stochastic depth \cite{Huang2016} enable the seemingly contradictory setup to train short networks and use deep networks at test time. This approach complements the recent success of residual networks. It reduces training time substantially and improves the test error significantly on almost all datasets. Recent evidence also indicates that CNNs could benefit from an interface to explicitly constructed memory mechanisms interacting with a CNN feature processing hierarchy. Correspondingly, the convolutional residual memory network \cite{Moniz2016} was proposed as a memory mechanism which enhances CNN architecture based on augmenting convolutional residual networks with a Long Short-Term Memory (LSTM) \cite{Hochreiter1997} mechanism. Weight normalization \cite{Salimans2016} was reported to be better suited for recurrent models such as LSTMs compared to batch normalization. It improves the conditioning of the optimization problem and speeds up convergence of stochastic gradient descent without introducing any dependencies between the examples in a mini-batch. Similarly, layer normalization \cite{Ba2016} normalizes across the inputs on a layer-by-layer basis at each time step. This stabilizes the dynamics of the hidden layers in the network and accelerates training, without the limitation of being tied to a batched implementation.

In this work, we develop a novel recurrent framework based on Recurrent Highway Networks (RHNs) for sequence modeling using batch normalization.  We explore the differences of several state-of-the-art techniques in terms of gradient control in data propagation within recurrent networks and compare the performance between them. The proposed technique relaxes the constraint in RHNs such that they have a better chance to avoid the gradient from vanishing or exploding by normalizing the recurrent transition units in highway layers. 

Section \ref{sec:relat} reviews some related techniques on gradient control in deep RNNs. Section \ref{sec:propo} presents the proposed sequence modeling framework and the details of the key component of the framework. Section \ref{sec:exper} discusses the experimental setup, performance evaluations, and experimental results. Finally, some concluding remarks are presented in Section \ref{sec:concl}.

\section{Related Work}
\label{sec:relat}

Much theoretical and empirical evidence indicates that the depth of neural networks plays an important role as a powerful machine learning paradigm. Deep CNNs have been proven successful on modern computer vision tasks, as illustrated in \cite{He2016}. However, increasing depth in RNNs which are deep in time domain typically does not take advantage of depth since the state update modeled by certain internal function mapping in modern RNNs is usually represented by non-linear activations \cite{Pascanu2013}. 

Recently, researchers made great efforts on gradient control. The highway layers \cite{Srivastava2015}, based on the LSTM unit, relax the limitation of training deep RNNs. Specifically, a highway network additionally defines two nonlinear transforms- the transform gate and carry gate.  These gates express how much of the output is produced by transforming the input and carrying it, respectively. By coupling the transform gate and carrying gate, a highway layer can smoothly vary its behavior between that of a plain layer and that of a layer which simply passes its inputs through. Due to this gating mechanism, a neural network can have paths along which information can flow across several layers without attenuation. Thus, highway networks, even with hundreds of layers, can be trained directly using stochastic gradient descent.  These networks, when used with a variety of activation functions, have been shown to avoid the vanishing or exploding gradient problem. Highway layers have achieved success in the fields of speech recognition \cite{Zhang2016} and language modeling \cite{Kim2015}.

Based on the insights of highway layers, Zilly \textit{et al.} \cite{Zilly2016} introduced Recurrent Highway Networks (RHNs) that have long credit assignment paths, not just in time, but also long in space (per time step). By replacing the LSTM cell in the recurrent loop, the RHN layer instead stacks the highway layers inside the recurrent units. By increasing recurrence depth, additional non-linearity strengthens the ability of the recurrent network without slowing down the processing. Compared to regular RNNs, RHNs provide more versatile ways to deal with data flow in terms of transforming and carrying information. It has been theoretically proven that coupling a carrying and transforming gate effectively controls the gradient.  However, such a constraint may limit the power of the network to some extent. In the next sections, we focus on this problem and propose a new scheme which relaxes the constraint in RHNs, by incorporating batch normalization.  Our method simultaneously improves network performance while avoiding the vanishing and exploding gradient problem.

\section{Proposed Framework}
\label{sec:propo}

A plain RNN consists of $L$ layers and $T$ time states. In general, each node in the layer $l\in\{1,2,...,L\}$ and time state $t\in\{1,2,...,T\}$ takes input $\textbf{x}_l^t$ and output $\textbf{h}_l^t$, respectively, with a non-linear transformation $H$. Omitting the bias term for simplicity, the output can be represented as 
\begin{align} \label{eq:1}
\textbf{h} = H(\textbf{x},\textbf{W}_H)
\end{align}
where the non-linear activation $H$ is typically specified by hyperbolic tangent function $tanh$, and $W_H$ is the associated weight matrix. In highway networks \cite{Srivastava2015}, the training process is facilitated by using adaptive computation. The additional defined transform gate and carry gate determine how much information is transformed and carried to the output, \textit{i.e.}, 
\begin{align} 
\label{eq:2} \textbf{t} =& T(\textbf{x},\textbf{W}_T) \\
\label{eq:3} \textbf{c} =& C(\textbf{x},\textbf{W}_C)
\end{align}
where $t,c$ are the output of the transform and carry gate respectively, $T,C$ are defined as a sigmoid function $\sigma(x) = 1/(1+e^{-x})$, and $\textbf{W}_T,\textbf{W}_C$ are corresponding weights. The RHN layer with recurrence depth $D$ is defined as 
\begin{align}
\textbf{s}_d^t=\textbf{h}_d^t \odot \textbf{t}_d^t + \textbf{s}_{d-1}^t \odot \textbf{c}_d^t
\end{align}
where $\odot$ implies the element-wise product.

RHN uses highway layers instead of LSTM units in regular recurrent networks as shown in the dotted box in Figure \ref{fig:archi}. Note that each recurrent loop takes the output of the last recurrent unit in the previous loop ($\textbf{s}_{D-1}^t$) as input, and the time-varying data $\textbf{x}^t$ is only fed into the recurrent loop to the recurrence depth, $d=1$. According to Ger\v{s}gorin circle theorem \cite{Gerschgorin1931}, all eigenvalues of the temporal Jacobian are preferably set to $1$ across time steps in order to keep the gradient flow steady. In this case, the Ger\v{s}gorin circle radius is reduced to 0 and each diagonal entry of temporal Jacobian is set to $1$. Zilly \textit{et al.} \cite{Zilly2016} states that it can be accomplished by the coupling the carry gate to the transform gate by setting $C=1-T$, as a constraint, in order to prevent an unbounded ``blow-up'' of state values which leads to more stable training. However, this constraint may limit the ability of the gates to freely learn parameter values and imposes a modeling bias which may be suboptimal for certain tasks \cite{Zaremba2015,Greff2016}. 

Figure \ref{fig:archi} shows the layout of the proposed recurrent network architecture. Because of its ability to control the gradient during back propagation, we incorporate batch normalization to the inputs of each recurrent loop. This allows us to relax the $C=1-T$ constraint, while simultaneously making gradients less prone to vanishing or exploding. Specifically, in batch normalization, the mean and variance are extracted across each channel and spatial locations. Each individual in the batch is normalized by subtracting the mean value and dividing by variance, and the data are recovered by shifting and scaling the normalized value during training.

\section{Experiments}
\label{sec:exper}

Experiments are performed on an image captioning task. The first part of this section describes the implementation details and experimental setup. Next, the evaluation and analysis of the proposed framework are discussed from different perspectives.

\subsection{Experimental Setup}
\label{ssec:exper_setup}

\hspace{15pt} \textbf{Dataset} The evaluation is carried out on the popular MSCOCO captioning dataset \cite{Lin2014}. This dataset contains $\sim$80k training images, $\sim$40k validation images and $\sim$40k test images. Note that ground truth captions are only available for training and validation sets. In order to efficiently use the available data, the validation set is split into three parts: 85\% of the images are merged into the training set, 10\% are used for testing, and the remaining 5\% are used as a validation set for hyperparamter tuning. All the experimental results are evaluated using the MSCOCO caption evaluation server \cite{Capeval2015}.

\textbf{Metrics} The following metrics are employed for evaluation:
1) \textit{BLEU} \cite{Papineni2002} is a metric for precision of word $n$-grams between predicted and ground truth sentences; 2) \textit{ROUGE-L} \cite{Lin2004} takes into account sentence level structure similarity naturally and identifies the longest co-occurring sequence in $n$-grams automatically; 3) \textit{METEOR} \cite{Banerjee2005} was designed to fix some of the problems found in the more popular BLEU metric, and also produce good correlation with human judgment at the sentence or segment level. It has several features not found in other metrics, such as stemming and synonymy matching, along with the standard exact word matching; and 4) \textit{CIDEr} \cite{Vedantam2015} computes the average cosine similarity between $n$-grams found in the generated caption and those found in reference sentences, weighting them using TF-IDF.  METEOR is more semantically preferred than BLEU and ROUGE-L \cite{Baraldi2016}. 

\textbf{Training Details} In the training phase, we add the \textless START\textgreater \  token at the beginning of the sentence and the \textless END\textgreater \ token at the end of the sentence so that the model can generate captions of varying lengths. In inference mode, the caption generation is started with \textless START\textgreater \ and the word combination with highest probability will be selected. The word embedding size and number of RHN neurons per layer are empirically set to $512$. Based on empirical results, we adopt the recurrence depth $D=3$. Stochastic gradient descent is employed for optimization, where the initial learning rate and decay factor are set to $0.1$ and $0.5$, respectively, and the learning rate decays exponentially every $8$ epochs. The initial time state vector is extracted from the Inception\_v3 model \cite{Szegedy2016} and all the other weight matrices are initialized with a random uniform distribution. The training process minimizes a softmax loss function. The proposed network is implemented using TensorFlow \cite{Tensorflow2015} and trained on a server with dual GeForce GTX 1080 graphics cards.

\subsection{Image Captioning Results}
\label{ssec:experi_resul}

We evaluated the proposed model on MSCOCO image captioning dataset. The results are reported in Table \ref{tab: mscoco_metric}. To make a fair comparison, we extract an image feature vector as initialization of the hidden state using the same Inception\_v3 model \cite{Szegedy2016}, and lock the parameters in it (without fine-tuning) in all test models. We compared three test models: LSTM denotes the im2txt model using regular LSTM cells implemented by \cite{Vinyals2016}; RHN denotes the image captioning generation performed by original RHNs \cite{Zilly2016}; and BN\_RHN is the proposed method with batch normalization instead of the $C=1-T$ constraint in RHN cell. The results show that the BN\_RHN is the best performing model. METEOR and CIDEr are generally considered the most robust scores for captioning. The higher BLEU-4 and METEOR scores, due to fluency of language in the image captions, can be attributed to the RHN depth. More depth increases the complexity that helps learn the grammatical rules and language semantics. The LSTM employs a mechanism with input, output, and forget gates to generate complex captions. Our model shows better performance than LSTM, which may indicate that simplifying the gate mechanism and increasing depth do not affect performance for image captioning. The test model with RHN cells benefits from having less parameters during training, and good gradient control, in a simple way. Our BN\_RHN achieves better result than original RHN, because the gate value model biases are more flexible, and batch normalization guarantees the steady gradient flow in back propagation.

\begin{table}[!htb] \centering
    \caption{Evaluation metrics on MSCOCO dataset. LSTM: regular RNN model with LSTM cell; RHN: model with original RHN cell; BN\_RHN: the proposed model with RHN constrain relaxed and batch normalization applied instead.}
    \label{tab: mscoco_metric}
    \begin{tabular}{l|ccc}
        \hline
        Model & LSTM &   RHN &  BN\_RHN \\
        \hline
        BLEU-1 & 0.706 & 0.688 & 0.710 \\
        BLEU-2 & 0.533 & 0.512 & 0.541 \\
        BLEU-3 & 0.397 & 0.377 & 0.408 \\
        BLEU-4 & 0.298 & 0.281 & 0.311 \\
        ROUGE-L & 0.524 & 0.511 & 0.533 \\
        METEOR & 0.248 & 0.241 & 0.254 \\
        CIDEr & 0.917 & 0.864 & 0.955 \\
        \hline        
    \end{tabular}
\end{table}

\begin{figure}[htb]
  \centering
  \centerline{\includegraphics[width=.39\textwidth, height=.3\textwidth,trim=3 2 0 0,clip]{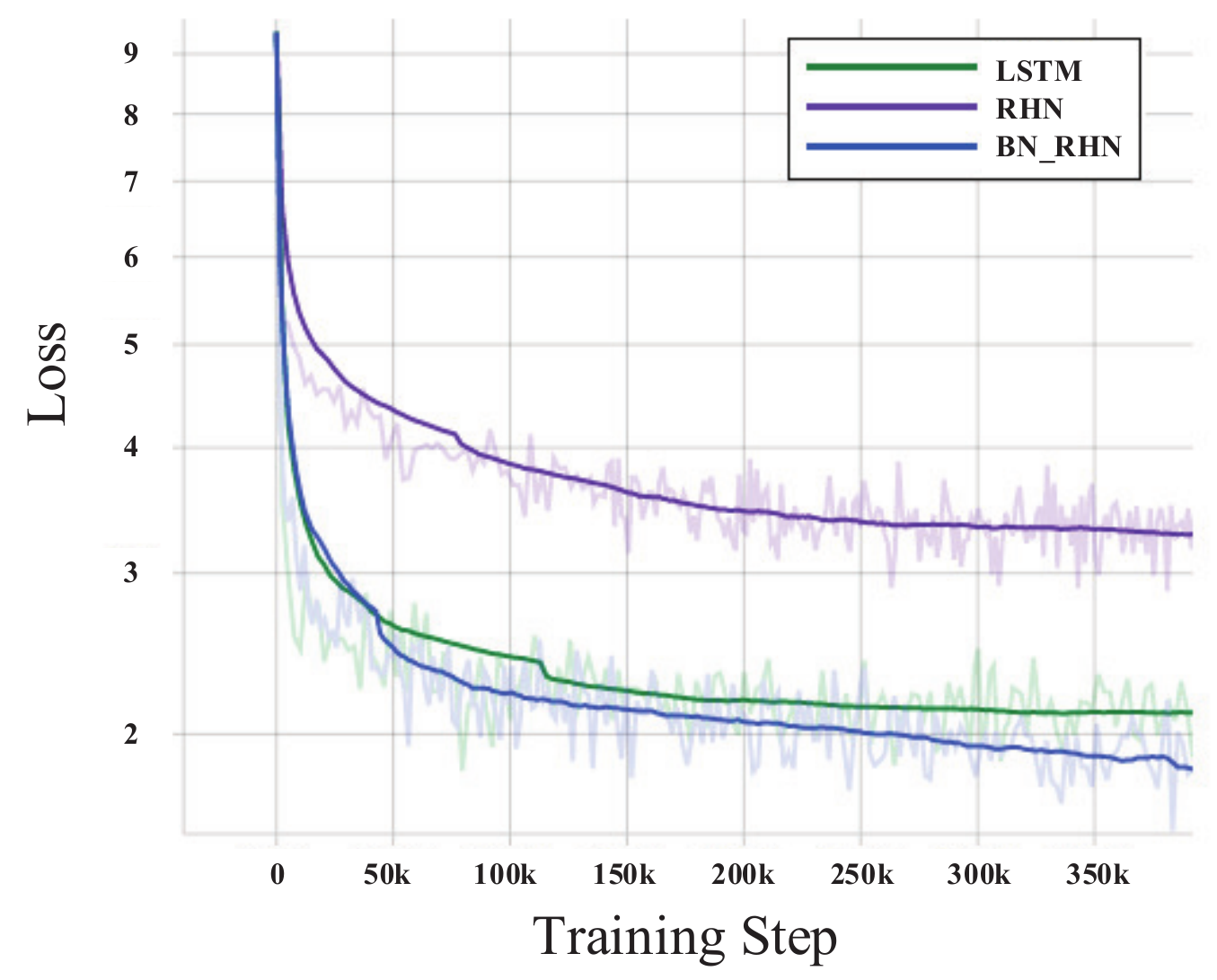}}
\caption{The total loss change vs. training steps. All dark curves are smoothed by a factor of 0.8. The light curves are not smoothed.}
\label{fig:loss}
\end{figure}

We additionally compare our model based on the speed of convergence. Figure \ref{fig:loss} shows the loss change during training. The BN\_RHN model achieves the steady loss fastest among all three models. It turns out that adding batch normalization allows a more aggressive learning rate and achieves faster convergence. It is worth mentioning that during back propagation in the original LSTM and RHN models, we have to adopt a gradient norm clipping strategy to deal with exploding gradients and a soft constraint for the vanishing gradients problem to generate reasonable captions. For BN\_RHN, this restriction can be relaxed. This confirms that the proposed model is effective on gradient control, as presupposed in Section \ref{sec:propo}.

\begin{figure}[htb]
\footnotesize
\begin{minipage}[b]{.48\linewidth}
  \centering
  \centerline{\includegraphics[width=3.0cm, height=3.0cm]{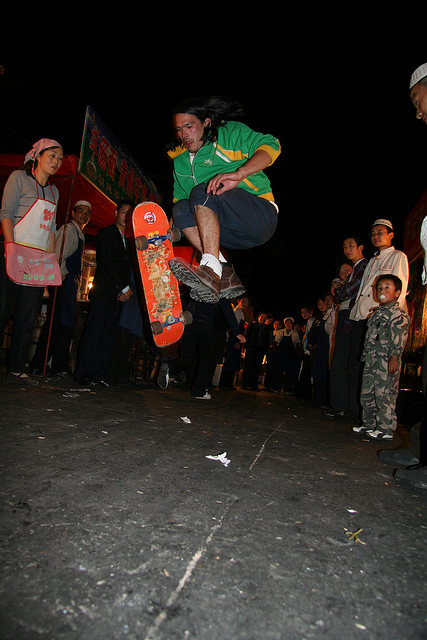}}
  \raggedright
  (\textbf{LSTM}) a group of people standing around a parking meter .
  (\textbf{RHN}) a group of people standing next to each other .
  (\textbf{BN\_RHN}) a young man riding a skateboard down a street .
  (\textbf{G.T.}) a person is doing a trick on a skateboard
\end{minipage}
\hfill
\begin{minipage}[b]{0.48\linewidth}
  \centering
  \centerline{\includegraphics[width=3.0cm, height=3.0cm]{COCO_val2014_000000305634.jpg}}
  \raggedright
  (\textbf{LSTM}) a red stop sign sitting on top of a metal pole . 
  (\textbf{RHN}) a red stop sign sitting on the side of a road .
  (\textbf{BN\_RHN}) a stop sign with a street sign attached to it .
  (\textbf{G.T.}) Street corner signs above a red stop sign.
\end{minipage}
\begin{minipage}[b]{.48\linewidth}
  \centering
  \centerline{\includegraphics[width=4.0cm, height=3.0cm]{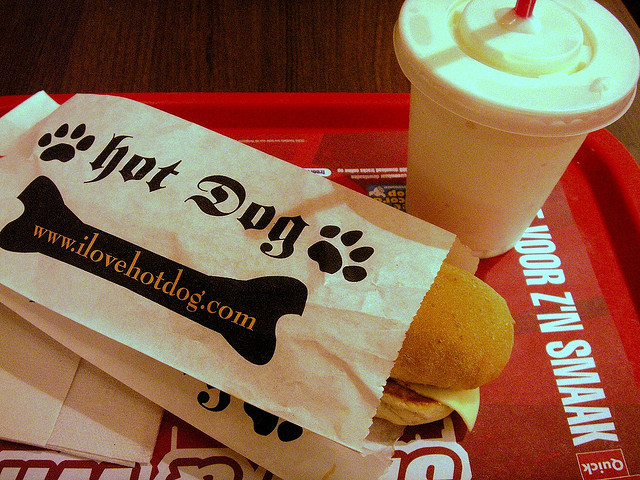}}
  \raggedright
  (\textbf{LSTM}) a box with a donut and a cup of coffee .
  (\textbf{RHN}) a birthday cake with a picture of a dog on it .
  (\textbf{BN\_RHN})  a plate with a doughnut and a cup of coffee . 
  (\textbf{G.T.}) A bag with a hot dog inside of it. \\
  \vphantom{placeholder}
\end{minipage}
\hfill
\begin{minipage}[b]{0.48\linewidth}
  \centering
  \centerline{\includegraphics[width=4.0cm, height=3.0cm]{COCO_val2014_000000576015.jpg}}
  \raggedright
  (\textbf{LSTM}) a rear view mirror of a car in the side view mirror .
  (\textbf{RHN}) a rear view mirror on the side of a car .
  (\textbf{BN\_RHN}) a rear view mirror with a dog in the side mirror .
  (\textbf{G.T.}) A guy takes a picture of his car's rear view mirror.
\end{minipage}
\vspace{-.1in}
\caption{Example results on MSCOCO captioning dataset. Top-left, top-right and bottom-left are positive examples; bottom-right is a negative example.}
\label{fig:example}
\end{figure}

In Figure \ref{fig:example}, we list a few representative examples of the descriptions generated by the proposed model, compared with the captions obtained by the original LSTM and RHN model. It is clear that the overall quality of the captions generated by our model have improved significantly compared to RHN model. We notice that BN\_RHN model describes the object in the image accurately and can generate better descriptions of the image even for very complex images. Additionally, the captions generated by the proposed model have better grammar and language semantics due to the increased depth of recurrent network.

\vspace{-.15in}
\section{Conclusions}
\label{sec:concl}

We introduce a novel recurrent neural network model that is based on batch normalization and recurrent highway networks. The analyses provide insight into the ability of the batch normalized recurrent highway model to dynamically control the gradient flow across time steps. Additionally, this model takes advantages of faster convergence compared to the original RHN, and keeps the feature of increasing depth in the recurrent transitions while retaining the ease of training. Experimental results on image captioning task reveals that our proposed model achieves high METEOR and BLEU scores compared to previous models on a modern dataset. 

\section*{\normalsize Acknowledgement}
\label{sec:ackno}
\vspace{-.1in}

\setstretch{0.5}{\small
The authors would like to express their gratitude to Kodak Alaris for their generous support of this research. They would also like to thank the Empire State Development’s Division of Science, Technology and Innovation (NYSTAR) and the Center for Emerging and Innovative Sciences, administered by the University of Rochester, for providing computation resources used in this research.
}

\setstretch{1.0}


\newpage
\bibliographystyle{IEEEbib}
\bibliography{strings,refs}

\end{document}